\documentclass{article}

\usepackage{arxiv}

\usepackage[utf8]{inputenc} 
\usepackage[T1]{fontenc}    
\usepackage{hyperref}       
\usepackage{url}            
\usepackage{booktabs}       
\usepackage{amsfonts}       
\usepackage{nicefrac}       
\usepackage{microtype}      
\usepackage{lipsum}
\usepackage{graphicx}

\usepackage[dvipsnames]{xcolor}
\usepackage{amsmath}
\usepackage{listings}
\usepackage{standalone}
\usepackage{tikz}
\usepackage{array}
\usepackage{tabularx}
\usepackage{float}

\hypersetup{
    colorlinks=true,
    linkcolor=blue,
}

\definecolor{codegreen}{rgb}{0,0.6,0}
\definecolor{codegray}{rgb}{0.5,0.5,0.5}
\definecolor{codepurple}{rgb}{0.58,0,0.82}
\definecolor{backcolour}{rgb}{0.99,0.99,0.96}

\lstdefinestyle{codeStyle}{
    backgroundcolor=\color{backcolour},   
    commentstyle=\color{codegreen},
    keywordstyle=\color{magenta},
    numberstyle=\tiny\color{codegray},
    stringstyle=\color{codepurple},
    basicstyle=\ttfamily\scriptsize,
    breakatwhitespace=false,         
    breaklines=true,                 
    captionpos=b,                    
    keepspaces=true,                 
    numbers=left,                    
    numbersep=3pt,                  
    showspaces=false,                
    showstringspaces=false,
    showtabs=false,                  
    tabsize=2
}
\graphicspath{ {./images/} }

\title{Domain-specific \texttt{ReAct} for physics-integrated iterative modeling: A case study of LLM agents for gas path analysis of gas turbines}

\author{
 Tao Song \\
  Department of Energy and Power Engineering\\
  Tsinghua University\\
  Beijing, China \\
  \texttt{songtao21@mails.tsinghua.edu.cn} \\
   \And
  Yuwei Fan \\
  Department of Energy and Power Engineering\\
  Tsinghua University\\
  Beijing, China \\
  \texttt{fan-yw22@mails.tsinghua.edu.cn} \\
  \And
  Chenlong Feng \\
  Department of Energy and Power Engineering\\
  Tsinghua University\\
  Beijing, China \\
  \texttt{fengcl23@mails.tsinghua.edu.cn} \\
  \And
  Keyu Song \\
  Department of Energy and Power Engineering\\
  Tsinghua University\\
  Beijing, China \\
  \texttt{sky20@mails.tsinghua.edu.cn} \\
  \And
  Chao Liu\thanks{Corresponding author.} \\
  Department of Energy and Power Engineering\\
  Tsinghua University\\
  Beijing, China \\
  \texttt{cliu5@tsinghua.edu.cn} \\
  \And
  Dongxiang Jiang \\
  Department of Energy and Power Engineering\\
  Tsinghua University\\
  Beijing, China \\
  \texttt{jiangdx@tsinghua.edu.cn} \\
}

\begin{document}
\maketitle
\begin{abstract}
This study explores the application of large language models (LLMs) with callable tools in energy and power engineering domain, focusing on gas path analysis of gas turbines. 
We developed a dual-agent tool-calling process to integrate expert knowledge, predefined tools, and LLM reasoning. We evaluated various LLMs, including LLama3, Qwen1.5 and GPT. 
Smaller models struggled with tool usage and parameter extraction, while larger models demonstrated favorable capabilities. 
All models faced challenges with complex, multi-component problems. 
Based on the test results, we infer that LLMs with nearly 100 billion parameters could meet professional scenario requirements with fine-tuning and advanced prompt design.
Continued development are likely to enhance their accuracy and effectiveness, paving the way for more robust AI-driven solutions.
\end{abstract}

\keywords{Large language models \and \texttt{ReAct} prompting \and Gas turbine}

\section{Introduction}

Artificial intelligence (AI) has seen extensive application across various domains, including energy and power engineering, where it holds significant promise for enhancing efficiency, reliability, and sustainability. 
Within the field of numerical computation, AI techniques have been effectively utilized to partly replace conventional numerical methods across a wide range of energy systems, aiding in the design and optimization of critical processes, such as structural strength and heat transfer \cite{chenHeatSourceLayout2020,andres-perezApplicationSurrogateRegression2021,fanStructuralDynamicLoad2023}.
Similarly, in power equipment signal analysis and fault diagnosis, AI algorithms have facilitated anomaly detection and predictive maintenance, thereby reducing downtime and enhancing operational efficiency\cite{jinHybridModelLSTM2021,songHierarchicalSchemeRemaining2022,jinNovelIntegratedModeling2022}.
Despite these successes, conventional AI tools often lack the theoretical foundation necessary to comprehensively address the complexities of energy systems. The black-box nature of many AI models hinders their interpretability, raising concerns regarding their trustworthiness and accountability in critical energy infrastructure applications.

Recently, the advent of large language models (LLMs) like GPT\cite{brownLanguageModelsAre2020}, LLaMA\cite{touvronLLaMAOpenEfficient2023}, and GLM\cite{duGLMGeneralLanguage2022} has revolutionized artificial intelligence, particularly in natural language processing (NLP). 
These models excel in generating human-like text, understanding context, and performing complex tasks. 
Concurrently, enhancement methods such as retrieval-augmented generation (RAG)\cite{lewisRetrievalAugmentedGenerationKnowledgeIntensive2021}, agent with tools\cite{paranjapeARTAutomaticMultistep2023}, and advanced prompting strategies like Chain-of-Thought\cite{weiChainofThoughtPromptingElicits2023} and Tree-of-Thoughts\cite{yaoTreeThoughtsDeliberate2023} have been proposed. 
These advancements address some issues of hallucinations and unexplainability in LLMs, making them more robust and trustworthy for practical applications.

In this context, we explored the feasibility of integrating LLMs into a domain-specific application scenarios, i.e. energy and power science and engineering, by focusing on a classic issue in gas turbine operation and maintenance: gas path analysis. 
Our method involved crafting specific prompts and a dual-agent tool-calling process to guide LLMs in parsing tool parameters. 
We defined energy equations as tools for each component and employed \texttt{ReAct} chain thinking prompts\cite{yaoReActSynergizingReasoning2023} to guide LLM reasoning and solve performance parameters. 
Testing across different LLMs provided insights into their capabilities for understanding, thinking, and reasoning, offering valuable perspectives for further research and applications.

\section{Background}

\subsection{Large Language Models}

Large language models (LLMs) epitomize a notable stride in artificial intelligence, particularly within the realm of natural language processing, hinging on neural networks and the transformer architecture for text processing and generation. 
Neural networks, intricate arrangements of interconnected neurons, adeptly transform input data through learned weights, thereby enabling multifaceted representations. 
A seminal advancement in LLMs is the integration of attention mechanisms, pivotal for discerning specific elements within input data, capturing extensive contextual dependencies. 
Central to LLMs, the transformer architecture comprises encoder and decoder layers, with the former processing input text and the latter generating output, leveraging self-attention mechanisms to gauge word relevance and contextual nuances. 
Beyond their fundamental architecture, LLMs offer distinct advantages over traditional AI methods, including their aptitude for coherent text generation and proficiency in reasoning tasks, bolstered by extensive training data. 
Techniques such as attention visualization enhance interpretability, shedding light on the model's underlying processing mechanisms. 

Recent years have witnessed a proliferation of LLM implementations, each characterized by unique architectural nuances, training datasets, and methodologies, engendering discernible disparities in their natural language comprehension and logical reasoning proficiencies. 
Our research evaluates multiple large language models from various organizations, elucidating their nuanced capabilities in parsing prompts and executing logical reasoning tasks for fundamental thermodynamics engineering problems.

\subsection{LLM agent and \texttt{ReAct} prompts}

The outputs of large language models are probabilistic estimations of human language, inherently subject to error and potential inaccuracies. These models may exhibit hallucinatory tendencies, particularly when confronted with scenarios beyond their training data, posing challenges for accurate response generation. Complex tasks often surpass the capabilities of single-shot language responses from large language models. Consequently, integrating these models with memory, planning, and tool-based capabilities emerges as a viable strategy for addressing intricate problems, leveraging their strengths while mitigating inherent limitations.

Large language model agents, enriched with callable tools and guided by \texttt{ReAct} (Reasoning and Acting) strategies, constitute an advanced framework for tackling intricate reasoning tasks. \texttt{ReAct} prompts facilitate systematic problem decomposition, addressing common LLM issues like hallucinations and unexplainability by providing structured reasoning pathways. Integrated callable tools serve as domain-specific functions, encapsulating expert knowledge and computational methodologies. In our experiment, a gas turbine component solution process, aligned with thermodynamic principles, guides LLMs to adhere to scientific laws. By synergizing LLMs' language comprehension and contextual processing prowess with human-defined tools, these agents achieve a balance between certainty and adaptability in problem-solving. This approach enhances the reliability and interpretability of LLM agents, catering to the nuanced demands of fields like energy and power engineering, characterized by complex, context-dependent challenges.

\subsection{Gas turbine components and test data}

The gas turbine, operating on the Brayton cycle, converts gaseous fuel into mechanical energy efficiently through compression, heating, and expansion stages. 
Its components, including the compressor, combustion chamber, turbine, and nozzle, play vital roles in this thermodynamic process. 
The compressor elevates air pressure through adiabatic compression, crucial for overall turbine performance. 
In the combustion chamber, fuel-air mixture combustion significantly increases temperature and pressure, while the turbine extracts energy from the resulting gas flow to produce mechanical work. 
The nozzle accelerates exhaust gases, enhancing turbine efficiency. 

\begin{figure}
    \centering
        \begin{tikzpicture}[x=0.75pt,y=0.75pt,yscale=-1,xscale=1]
        
        \draw  [fill={rgb, 255:red, 98; green, 249; blue, 255 }  ,fill opacity=0.85 ] (128.5,216.21) -- (74.75,269.96) -- (75.25,83.25) -- (128.71,136.72) -- cycle ;
        \draw  [fill={rgb, 255:red, 255; green, 147; blue, 42 }  ,fill opacity=0.61 ] (258.78,136.72) -- (312.53,82.97) -- (312.03,269.68) -- (258.57,216.21) -- cycle ;
        \draw  [fill={rgb, 255:red, 255; green, 78; blue, 51 }  ,fill opacity=0.63 ] (128.5,136.72) -- (258.57,136.72) -- (258.57,216.21) -- (128.5,216.21) -- cycle ;
        \draw  [color={rgb, 255:red, 0; green, 0; blue, 0 }  ,draw opacity=1 ][fill={rgb, 255:red, 0; green, 0; blue, 0 }  ,fill opacity=1 ] (128.5,165.97) -- (258.57,165.97) -- (258.57,186.97) -- (128.5,186.97) -- cycle ;
        \draw  [fill={rgb, 255:red, 252; green, 255; blue, 98 }  ,fill opacity=0.85 ] (341,240.71) -- (312.03,269.68) -- (312.53,82.97) -- (341.34,111.78) -- cycle ;
        \draw    (75.25,83.25) -- (75.25,69.27) ;
        \draw [shift={(75.25,67.27)}, rotate = 90] [color={rgb, 255:red, 0; green, 0; blue, 0 }  ][line width=0.75]    (10.93,-3.29) .. controls (6.95,-1.4) and (3.31,-0.3) .. (0,0) .. controls (3.31,0.3) and (6.95,1.4) .. (10.93,3.29)   ;
        \draw    (128.71,136.72) -- (128.71,69.27) ;
        \draw [shift={(128.71,67.27)}, rotate = 90] [color={rgb, 255:red, 0; green, 0; blue, 0 }  ][line width=0.75]    (10.93,-3.29) .. controls (6.95,-1.4) and (3.31,-0.3) .. (0,0) .. controls (3.31,0.3) and (6.95,1.4) .. (10.93,3.29)   ;
        \draw    (258.57,216.21) -- (258.57,273.27) ;
        \draw [shift={(258.57,275.27)}, rotate = 270] [color={rgb, 255:red, 0; green, 0; blue, 0 }  ][line width=0.75]    (10.93,-3.29) .. controls (6.95,-1.4) and (3.31,-0.3) .. (0,0) .. controls (3.31,0.3) and (6.95,1.4) .. (10.93,3.29)   ;
        \draw    (312.53,82.97) -- (312.53,69.27) ;
        \draw [shift={(312.53,67.27)}, rotate = 90] [color={rgb, 255:red, 0; green, 0; blue, 0 }  ][line width=0.75]    (10.93,-3.29) .. controls (6.95,-1.4) and (3.31,-0.3) .. (0,0) .. controls (3.31,0.3) and (6.95,1.4) .. (10.93,3.29)   ;
        \draw    (341.34,111.78) -- (341.34,69.27) ;
        \draw [shift={(341.34,67.27)}, rotate = 90] [color={rgb, 255:red, 0; green, 0; blue, 0 }  ][line width=0.75]    (10.93,-3.29) .. controls (6.95,-1.4) and (3.31,-0.3) .. (0,0) .. controls (3.31,0.3) and (6.95,1.4) .. (10.93,3.29)   ;
        \draw    (153.25,136.72) -- (153.25,69.27) ;
        \draw [shift={(153.25,67.27)}, rotate = 90] [color={rgb, 255:red, 0; green, 0; blue, 0 }  ][line width=0.75]    (10.93,-3.29) .. controls (6.95,-1.4) and (3.31,-0.3) .. (0,0) .. controls (3.31,0.3) and (6.95,1.4) .. (10.93,3.29)   ;
        \draw    (128.5,216.21) -- (128.5,273.27) ;
        \draw [shift={(128.5,275.27)}, rotate = 270] [color={rgb, 255:red, 0; green, 0; blue, 0 }  ][line width=0.75]    (10.93,-3.29) .. controls (6.95,-1.4) and (3.31,-0.3) .. (0,0) .. controls (3.31,0.3) and (6.95,1.4) .. (10.93,3.29)   ;
        \draw    (341,240.71) -- (341,273.27) ;
        \draw [shift={(341,275.27)}, rotate = 270] [color={rgb, 255:red, 0; green, 0; blue, 0 }  ][line width=0.75]    (10.93,-3.29) .. controls (6.95,-1.4) and (3.31,-0.3) .. (0,0) .. controls (3.31,0.3) and (6.95,1.4) .. (10.93,3.29)   ;
        \draw  [dash pattern={on 4.5pt off 4.5pt}] (23,55.57) .. controls (23,46.42) and (30.42,39) .. (39.57,39) -- (401.64,39) .. controls (410.79,39) and (418.21,46.42) .. (418.21,55.57) -- (418.21,55.71) .. controls (418.21,64.87) and (410.79,72.29) .. (401.64,72.29) -- (39.57,72.29) .. controls (30.42,72.29) and (23,64.87) .. (23,55.71) -- cycle ;
        \draw  [dash pattern={on 4.5pt off 4.5pt}] (115.21,287.29) .. controls (115.21,279) and (121.93,272.29) .. (130.21,272.29) -- (369.71,272.29) .. controls (378,272.29) and (384.71,279) .. (384.71,287.29) -- (384.71,287.29) .. controls (384.71,295.57) and (378,302.29) .. (369.71,302.29) -- (130.21,302.29) .. controls (121.93,302.29) and (115.21,295.57) .. (115.21,287.29) -- cycle ;
        \draw  [line width=1.5]  (434,63) .. controls (434,58.58) and (437.58,55) .. (442,55) -- (686.07,55) .. controls (690.49,55) and (694.07,58.58) .. (694.07,63) -- (694.07,87) .. controls (694.07,91.42) and (690.49,95) .. (686.07,95) -- (442,95) .. controls (437.58,95) and (434,91.42) .. (434,87) -- cycle ;
        \draw  [line width=1.5]  (434,127) .. controls (434,122.58) and (437.58,119) .. (442,119) -- (686.07,119) .. controls (690.49,119) and (694.07,122.58) .. (694.07,127) -- (694.07,151) .. controls (694.07,155.42) and (690.49,159) .. (686.07,159) -- (442,159) .. controls (437.58,159) and (434,155.42) .. (434,151) -- cycle ;
        \draw  [line width=1.5]  (434,188.04) .. controls (434,182.54) and (438.46,178.07) .. (443.97,178.07) -- (684.1,178.07) .. controls (689.61,178.07) and (694.07,182.54) .. (694.07,188.04) -- (694.07,217.96) .. controls (694.07,223.46) and (689.61,227.93) .. (684.1,227.93) -- (443.97,227.93) .. controls (438.46,227.93) and (434,223.46) .. (434,217.96) -- cycle ;
        \draw  [line width=1.5]  (434,255) .. controls (434,250.58) and (437.58,247) .. (442,247) -- (686.07,247) .. controls (690.49,247) and (694.07,250.58) .. (694.07,255) -- (694.07,279) .. controls (694.07,283.42) and (690.49,287) .. (686.07,287) -- (442,287) .. controls (437.58,287) and (434,283.42) .. (434,279) -- cycle ;
        \draw [line width=1.5]    (564.04,158.29) -- (564.04,174.29) ;
        \draw [shift={(564.04,177.29)}, rotate = 270] [color={rgb, 255:red, 0; green, 0; blue, 0 }  ][line width=1.5]    (8.53,-2.57) .. controls (5.42,-1.09) and (2.58,-0.23) .. (0,0) .. controls (2.58,0.23) and (5.42,1.09) .. (8.53,2.57)   ;
        \draw [line width=1.5]    (564.04,229.29) -- (564.04,243.29) ;
        \draw [shift={(564.04,246.29)}, rotate = 270] [color={rgb, 255:red, 0; green, 0; blue, 0 }  ][line width=1.5]    (8.53,-2.57) .. controls (5.42,-1.09) and (2.58,-0.23) .. (0,0) .. controls (2.58,0.23) and (5.42,1.09) .. (8.53,2.57)   ;

        \draw (564.04,75) node   [align=left] {Compressor: $\displaystyle T_{1} ,p_{1} ,T_{2} ,p_{2}\rightarrow \eta _{comp}$};
        \draw (564.04,139) node   [align=left] {Nozzle: $\displaystyle T_{4} ,p_{4} ,p_{5} ,A_{throat}\rightarrow \dot{m}_{gas}$};
        \draw (564.04,203) node   [align=left] {Combustion chamber: \\$\displaystyle T_{2} ,p_{2} ,\dot{m}_{gas} ,\dot{m}_{fuel}\rightarrow T_{3} ,p_{3}$};
        \draw (564.04,267) node   [align=left] {Turbine: $\displaystyle \ T_{3} ,p_{3} ,T_{4} ,p_{4}\rightarrow \eta _{turb}$};
        \draw (98,176.61) node  [rotate=-90] [align=left] {Compressor};
        \draw (193.54,176.47) node  [color={rgb, 255:red, 255; green, 255; blue, 255 }  ,opacity=1 ] [align=left] {Shaft};
        \draw (193.54,152.47) node  [font=\footnotesize,color={rgb, 255:red, 0; green, 0; blue, 0 }  ,opacity=1 ] [align=left] {Combustion chamber};
        \draw (283.28,176.32) node  [rotate=-90] [align=left] {Turbine};
        \draw (73.25,63.87) node [anchor=south east] [inner sep=0.75pt]    {$T_{1} ,p_{1}$};
        \draw (126.71,63.87) node [anchor=south east] [inner sep=0.75pt]    {$T_{2} ,p_{2}$};
        \draw (310.53,63.87) node [anchor=south east] [inner sep=0.75pt]    {$T_{4} ,p_{4}$};
        \draw (155.25,63.87) node [anchor=south west] [inner sep=0.75pt]    {$\dot{m}_{fuel}$};
        \draw (325.28,176.32) node  [rotate=-90] [align=left] {Nozzle};
        \draw (343.34,63.87) node [anchor=south west] [inner sep=0.75pt]    {$p_{5} ,A_{throat}$};
        \draw (256.57,278.67) node [anchor=north east] [inner sep=0.75pt]    {$T_{3} ,p_{3}$};
        \draw (130.5,278.67) node [anchor=north west][inner sep=0.75pt]    {$\dot{m}_{air}$};
        \draw (343,278.67) node [anchor=north west][inner sep=0.75pt]    {$\dot{m}_{gas}$};
        \draw (41.57,36) node [anchor=south west] [inner sep=0.75pt]  [font=\footnotesize,color={rgb, 255:red, 0; green, 0; blue, 0 }  ,opacity=1 ] [align=left] {Measured parameters};
        \draw (129.71,305.29) node [anchor=north west][inner sep=0.75pt]  [font=\footnotesize,color={rgb, 255:red, 0; green, 0; blue, 0 }  ,opacity=1 ] [align=left] {Parameters not measured};
    \end{tikzpicture}
    
    \caption{Schema of gas path analysis for gas turbines}
    \label{fig:gas-turb-calc}
\end{figure}
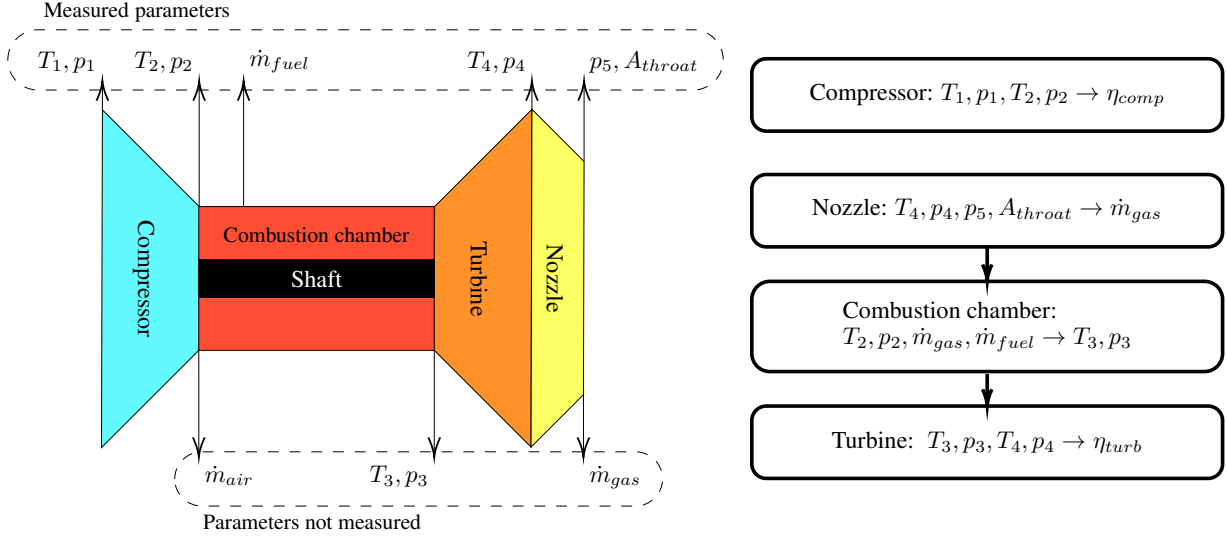

Calculating compressor and turbine adiabatic efficiency involves determining enthalpy changes across these components under adiabatic assumptions and comparing actual to theoretical enthalpy changes to gauge performance. 
Test conditions typically measure temperatures, pressures, and fuel flow rates, while complexities often exclude parameters such as combustion chamber outlet (turbine inlet) and air/gas flow rates.
Consequently, calculating turbine isentropic efficiency requires solving multiple components: determining gas flow rate through the nozzle, calculating turbine inlet temperature and pressure via enthalpy conservation in the combustion chamber, and finally computing the turbine's isentropic efficiency.
This process is illustrated in Figure \ref{fig:gas-turb-calc}.
This article aims to enable the large language model to comprehend and execute this calculation process.

\section{Methods}

\subsection{LLMs models}

Given the varying capabilities of different large language models, we selected models from various institutions to enable the calculation of gas turbine performance parameters and assess their application potential in energy and power engineering. 
The details of these models are presented in Table \ref{tab:llms}.

\begin{table}[htbp]
    \centering
    \footnotesize
    \caption{LLMs used in experiment}
        \begin{tabular}{m{5em}m{4.145em}m{4.145em}m{8.4em}m{8em}m{5.5em}m{5.44em}}
        \toprule 
        Models & Number of Params & Model Creators & Publication Date & Access Method & Native Tools Support & Quantization  Method \\
        \midrule
        LLama3\cite{touvronLLaMAOpenEfficient2023}     & 8B         & Meta       & April 18, 2024 & Local Deployment & No         & FP16 \\
        LLama3     & 70B        & Meta       & April 18, 2024 & Local Deployment & No         & 4Bit \\
        Qwen1.5\cite{baiQwenTechnicalReport2023}    & 32B        & Alibaba    & February 5, 2024 & Local Deployment & No         & 4Bit \\
        Qwen1.5    & 72B        & Alibaba    & February 5, 2024 & Local Deployment & No         & 4Bit \\
        GPT3.5\cite{brownLanguageModelsAre2020}     & -          & OpenAI     & November 30, 2022 & API        & Yes        & - \\
        GPT4o     & -          & OpenAI     & May 13, 2024 & Web        & Yes        & - \\
        GLM4\cite{duGLMGeneralLanguage2022}       & -          & ZhipuAI    & January 16, 2024 & Web \& API        & Yes        & - \\
        \bottomrule
        \end{tabular}%
    \label{tab:llms}%
\end{table}%

Due to the commercial sensitivity of data in the energy and power engineering industry, we focused on testing multiple large language models that can be locally executed on consumer-grade graphics cards. 
The models listed in Table \ref{tab:llms} were deployed on a platform equipped with two NVIDIA RTX 4090 GPUs (totaling 48 GB of video memory), with quantization applied to meet the memory constraints. 
Additionally, we tested the process described in this article using business models accessed through APIs to facilitate a comparative analysis of the results.

\subsection{Dual-agent tool-calling process}

\begin{figure}
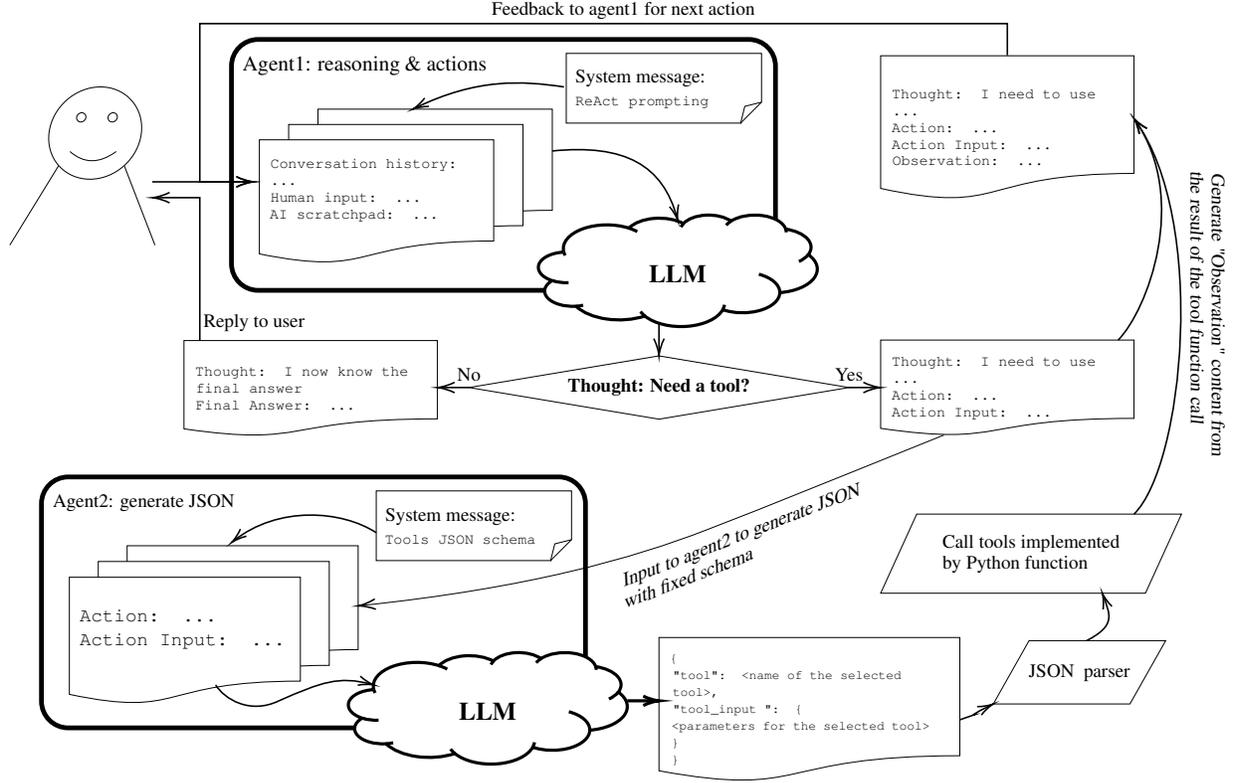

    \centering
    \includestandalone[width=0.98\textwidth]{dual-agent-process}
    \caption{Dual-agent tool-calling process}
    \label{fig:dual-agent-process}
\end{figure}

Given that the aforementioned large language models do not all natively support tool calling, this article employs a general dual-agent tool-calling process to construct a unified large language model agent. 
As illustrated in Figure \ref{fig:dual-agent-process}, the first agent, prompted by the \texttt{ReAct} framework, outputs "Thought, Action, Action input" logically, enabling it to select and plan subsequent actions. 
When tool usage is required, the Action and Action input content generated by the first agent are submitted to the second agent. 
The second agent is tasked with qualifying the tool and extracting specific information in JSON format. 
This JSON information is then parsed into Python variables, and the preset tool code is executed to complete the calculation. 
The result from the second tool is returned as an "Observation" content to the first agent, which continues the "Thought, Action, Action input" process until it outputs the "Thought, Final Answer".
Additionally, the first agent's prompt includes several sentences on the fundamentals of gas turbines. 
The prompt words for both agents are provided in the appendix \ref{sec:llm-prompts} at the end of this article.

\subsection{Gas path analysis solver tools}

To ensure the outputs generated by the LLMs adhere to the fundamental laws of thermodynamics and to assist the agents in calculating the isentropic efficiency of the gas turbine compressor and turbine, we have predefined parameters and developed four tools for the gas turbine's compressor, combustion chamber, turbine, and nozzle. 
These tools operate according to the following principles:

\begin{itemize}
    \item Compressor
    \[ \eta_{\text{comp}} = \frac{\Delta H_\text{ideal}}{\Delta H_\text{real}} \approx \frac{\left( \frac{P_\text{out}}{P_\text{in}} \right) ^ {\frac{\gamma - 1 }{\gamma}} - 1}{\frac{T_\text{out}}{T_\text{in}} - 1} \]
    \item Combustion Chamber
    \[ \dot{m}_\text{in} h_\text{in,air} + \dot{m}_\text{fuel} q_\text{fuel} = \dot{m}_\text{out} h_\text{out,gas} \Rightarrow  T_\text{out} \approx T_\text{in} + \frac{q_\text{fuel} \dot{m}_\text{fuel}}{c_{p,\text{air}} \dot{m}_\text{out}} \]
    \item Turbine
    \[ \eta_{\text{turb}} = \frac{\Delta H_\text{real}}{\Delta H_\text{ideal}} \approx \frac{1-\frac{T_\text{out}}{T_\text{in}}}{1 - \left( \frac{P_\text{out}}{P_\text{in}} \right) ^ {\frac{\gamma - 1 }{\gamma}}} \]
    \item Nozzle
    \scriptsize \[ \text{Mach}_{\text{out}} = \min\left(\sqrt{\left(\left(\frac{P_{\text{in}}}{P_{\text{out}}}\right)^{\frac{\gamma - 1}{\gamma}} - 1\right) \cdot \frac{2}{\gamma - 1}}, 1\right) \Rightarrow \dot{m}_{\text{out}} \approx P_{\text{in}} A_{\text{throat}} \sqrt{\frac{\gamma}{R \cdot T_{\text{in}}}} \text{Mach}_{\text{out}} \cdot \left(1 + \frac{\gamma - 1}{2} \text{Mach}_{\text{out}}^2\right)^{-\frac{\gamma + 1}{2(\gamma - 1)}} \]
\end{itemize}

To streamline the calculation process, various approximations were made, particularly concerning gas thermophysical properties and flow conservation. 
Since the internal calculations of these tools are not visible to the large language model, these simplifications do not impact the experimental evaluation of the model's agent capabilities. 
In practical engineering applications, these tools can be substituted with any calculation process necessary to meet modeling accuracy requirements. 
The functions defining these four calculation tools are detailed in the appendix \ref{sec:tools-code} at the end of this article.


\section{Results and discussion}

In the initial phase, we evaluated the performance of LLMs in calculating single component without embedded tools. 
Detailed questions, can be found in the appendix \ref{sec:test-questions}, were posed to the models. The results, summarized in Table \ref{tab:native-llms-result}, revealed that locally deployed models (LLama3 and Qwen1.5) could only engage in text-based reasoning and preliminary calculations, resulting in incorrect outcomes. 
This highlights their limitation in performing accurate technical calculations through text alone.
Conversely, web-accessed models (ChatGPT4o and ChatGLM4) demonstrated better performance. 
They engaged in conceptual reasoning, made reasonable assumptions (e.g., ideal gas properties, fuel calorific value, etc.), and generated and executed Python code. 
ChatGPT4o accurately calculated all gas turbine components, while ChatGLM4 correctly calculated the combustion chamber and turbine but erred with the nozzle and compressor.

These results indicate that while LLMs have a foundational understanding of gas turbine concepts, their reasoning and calculation accuracy varies. 
Web-accessed models, which support Python script execution, can handle complex tasks better than locally deployed models, which are limited to text-based reasoning.

\begin{table}[htbp]
    \centering
    \caption{Reasoning and calculation results of single components by native LLMs}
    \footnotesize
    \begin{tabular}{lll}
        \toprule
        Models & Access Met. & Result \\
        \midrule
        LLama3-8B     & Local Depl. & Incorrect reasoning \& calculation through formulas expressed in textual language \\
        LLama3-70B     & Local Depl. & Incorrect reasoning \& calculation through formulas expressed in textual language \\
        Qwen1.5-32B    & Local Depl. & Incorrect reasoning \& calculation through formulas expressed in textual language \\
        Qwen1.5-72B    & Local Depl. & Incorrect reasoning \& calculation through formulas expressed in textual language \\
        \textbf{GPT4o}      & Web        & \textbf{All correct calculation through reasoning, generation \& execution of Python code} \\
        GLM4       & Web        & Partially incorrect calculation through reasoning, generation \& execution of Python code \\
        \bottomrule
    \end{tabular}%
    \label{tab:native-llms-result}%
\end{table}%

Most LLMs lack sufficient native knowledge of gas turbine components and cannot perform reasoning and calculations independently. 
To address this, we built a dual-agent system with predefined gas turbine calculation tools. After construction, we tested single-component calculation questions, summarized in Table \ref{tab:agent-tools-result}.
For the small model LLama3-8B, agent1 correctly generated "Thought, Action, Action input" response, but agent2 incorrectly produced tool call results in JSON when more than three tools were available. 
The Qwen1.5 models' agent1 functioned normally, but agent2's JSON strings had subtle errors, like added spaces in parameter names.
Conversely, LLama3-70B, GPT3.5 and GLM4 accurately completed tool calls for all four components and generated user-friendly replies, demonstrating its superior capabilities. 
Detailed responses from some agents are provided in the appendix \ref{sec:agent-response}.

\begin{table}[htbp]
    \centering
    \caption{Reasoning and calculation results of single components by LLM agent embedded with tools}
    \footnotesize
    \begin{tabular}{lll}
        \toprule
        Models & Access Met. & Result \\
        \midrule
        LLama3-8B  & Local Depl. & Correct Action info from agent1, incorrect tool call JSON from agent2 \\
        \textbf{LLama3-70B} & Local Depl. & \textbf{Correct results from both agents} \\
        Qwen1.5-32B & Local Depl. & Correct Action info from agent1, incorrect tool call JSON from agent2 \\
        Qwen1.5-72B & Local Depl. & Correct Action info from agent1, incorrect tool call JSON from agent2 \\
        \textbf{GPT3.5}     & API        & \textbf{Correct results from both agents}  \\
        \textbf{GLM4}       & API        & \textbf{Correct results from both agents} \\
        \bottomrule
    \end{tabular}%
    \label{tab:agent-tools-result}%
\end{table}%

After integrating multi-component calculation tools, the three LLM agents could reason and make tool call decisions based on user queries. 
We tested them on a more complex task: calculating the adiabatic efficiency of both the gas turbine's compressor and turbine using typical test data. 
Detailed questions are provided in the appendix \ref{sec:test-questions}.
To solve this, the LLM agent needed to sequentially call the tools according to the logic in Figure \ref{fig:gas-turb-calc}, selecting correct parameters from the input and output data. 
Initially, all three agents correctly used the measured inlet and outlet data of the compressor. 
However, they incorrectly used unmeasured turbine inlet temperature and pressure parameters, leading to wrong turbine efficiency results. 
This indicates the LLMs struggled with complex reasoning chains and lacked a deep understanding of gas turbine specifics.
With hints provided for solving sequence, GLM4 and GPT3.5 adjusted their order correctly but still selected incorrect tool parameters for the nozzle, resulting in errors. 
In contrast, LLama3-70B correctly executed the entire chain solution. Responses from LLM agents are included in the appendix \ref{sec:agent-response}.

\begin{table}[htbp]
    \centering
    \footnotesize
    \caption{Chain calculation results of multiple components by LLM agent with hints}
    \begin{tabular}{lll}
        \toprule
        Models & Access Met. & Result \\
        \midrule
        \textbf{LLama3-70B} & Local Depl. & \textbf{Correctly executed the entire chain solution} \\
        GPT3.5     & API        & Incorrectly selected parameters for calculating the nozzle \\
        GLM4       & API        & Incorrectly selected parameters for calculating the nozzle \\
        \bottomrule
    \end{tabular}%
    \label{tab:multi-tools-result}%
\end{table}%

The above results show that different LLMs exhibit varying capabilities. 
The state-of-the-art GPT4o model can independently generate and execute correct Python scripts for simple thermodynamic problems, such as gas turbine component calculations, highlighting the potential of tool-embedded LLM agents for scientific and engineering applications.
Our dual-agent tool-calling process enables tool use across various LLMs, regardless of native support, reducing illusion phenomena and integrating human knowledge and complex models. 
However, LLMs vary in tool selection and parameter extraction accuracy. 
Generally, larger, more recent models perform better. 
Despite advancements, current LLMs struggle with complex, multi-component gas turbine problems. 
With human knowledge planning, LLMs may eventually handle these challenges successfully.

\section{Conclusion}

Our study investigated the application of large language models (LLMs) with callable tools in the specialized domain of energy and power engineering, focusing specifically on gas turbine gas path analysis. 
We implemented a dual-agent tool-calling process designed to enhance LLM capabilities by integrating human knowledge, predefined tools, and LLM reasoning. 
This process was tested on several mainstream LLMs, including LLama3, Qwen1.5, and GPT, to evaluate their proficiency in tool parameter extraction, knowledge understanding, and reasoning.
Smaller and older models such as LLama3-8B and Qwen1.5 exhibited limitations in effectively utilizing tools and extracting parameters, underscoring their struggles with complex, multi-component problems. 
In contrast, more advanced models like GPT4o and LLama3-70B demonstrated superior capabilities, showing promise in handling intricate engineering tasks.
Despite these advancements, current LLMs still face challenges with highly complex, multi-component problems. 
However, LLM agents with tool calling ability, combined with human expertise, show significant potential for improving LLM performance in these intricate tasks. 
Continued development, fine-tuning, and advanced prompt engineering are likely to enhance their accuracy and effectiveness, enabling more sophisticated problem-solving in specialized domains.

Based on the test results of these models, we can reasonably anticipate that with fine-tuning using professional knowledge and advanced prompting design, LLMs with nearly 100 billion parameters could meet the reasoning requirements of professional scenarios. 
Our findings lay the groundwork for future research and development in the application of LLMs in specialized engineering fields, paving the way for more robust and capable AI-driven solutions.



\begin{thebibliography}{10}

    \bibitem{chenHeatSourceLayout2020}
    Xiaoqian Chen, Xianqi Chen, Weien Zhou, Jun Zhang, and Wen Yao.
    \newblock The heat source layout optimization using deep learning surrogate
      modeling.
    \newblock {\em Structural and Multidisciplinary Optimization},
      62(6):3127--3148, December 2020.
    
    \bibitem{andres-perezApplicationSurrogateRegression2021}
    Esther {Andr{\'e}s-P{\'e}rez} and Carlos {Paulete-Peri{\'a}{\~n}ez}.
    \newblock On the application of surrogate regression models for aerodynamic
      coefficient prediction.
    \newblock {\em Complex \& Intelligent Systems}, 7(4):1991--2021, August 2021.
    
    \bibitem{fanStructuralDynamicLoad2023}
    Yuwei Fan, Tengbo Guo, Zhe Li, Liangyou Hong, Chao Liu, and Dongxiang Jiang.
    \newblock {Structural Dynamic Load Prediction Method Based on Long Short-term
      Memory Network}.
    \newblock {\em Chinese Journal of Ship Research}, 2023.
    
    \bibitem{jinHybridModelLSTM2021}
    Yunfeng Jin, Chao Liu, Xin Tian, Haizhou Huang, Gaofeng Deng, Yunlong Guan, and
      Dongxiang Jiang.
    \newblock A hybrid model of {{LSTM}} neural networks with a thermodynamic model
      for condition-based maintenance of compressor fouling.
    \newblock {\em Measurement Science and Technology}, 32(12):124007, September
      2021.
    
    \bibitem{songHierarchicalSchemeRemaining2022}
    Tao Song, Chao Liu, Rui Wu, Yunfeng Jin, and Dongxiang Jiang.
    \newblock A hierarchical scheme for remaining useful life prediction with long
      short-term memory networks.
    \newblock {\em Neurocomputing}, 487:22--33, May 2022.
    
    \bibitem{jinNovelIntegratedModeling2022}
    Yunfeng Jin, Chao Liu, Xin Tian, Haizhou Huang, Gaofeng Deng, Yunlong Guan, and
      Dongxiang Jiang.
    \newblock A novel integrated modeling approach for filter diagnosis in gas
      turbine air intake system.
    \newblock {\em Proceedings of the Institution of Mechanical Engineers, Part A:
      Journal of Power and Energy}, 236(3):435--449, May 2022.
    
    \bibitem{brownLanguageModelsAre2020}
    Tom~B. Brown, Benjamin Mann, Nick Ryder, Melanie Subbiah, Jared Kaplan,
      Prafulla Dhariwal, Arvind Neelakantan, Pranav Shyam, Girish Sastry, Amanda
      Askell, Sandhini Agarwal, Ariel {Herbert-Voss}, Gretchen Krueger, Tom
      Henighan, Rewon Child, Aditya Ramesh, Daniel~M. Ziegler, Jeffrey Wu, Clemens
      Winter, Christopher Hesse, Mark Chen, Eric Sigler, Mateusz Litwin, Scott
      Gray, Benjamin Chess, Jack Clark, Christopher Berner, Sam McCandlish, Alec
      Radford, Ilya Sutskever, and Dario Amodei.
    \newblock Language {{Models}} are {{Few-Shot Learners}}, July 2020, {\em arXiv preprint arXiv:2005.14165}.
    
    \bibitem{touvronLLaMAOpenEfficient2023}
    Hugo Touvron, Thibaut Lavril, Gautier Izacard, Xavier Martinet, Marie-Anne
      Lachaux, Timoth{\'e}e Lacroix, Baptiste Rozi{\`e}re, Naman Goyal, Eric
      Hambro, Faisal Azhar, Aurelien Rodriguez, Armand Joulin, Edouard Grave, and
      Guillaume Lample.
    \newblock {{LLaMA}}: {{Open}} and {{Efficient Foundation Language Models}},
      February 2023, {\em arXiv preprint arXiv:2302.13971}.
    
    \bibitem{duGLMGeneralLanguage2022}
    Zhengxiao Du, Yujie Qian, Xiao Liu, Ming Ding, Jiezhong Qiu, Zhilin Yang, and
      Jie Tang.
    \newblock {{GLM}}: {{General Language Model Pretraining}} with {{Autoregressive
      Blank Infilling}}, March 2022, {\em arXiv preprint arXiv:2103.10360}.
    
    \bibitem{lewisRetrievalAugmentedGenerationKnowledgeIntensive2021}
    Patrick Lewis, Ethan Perez, Aleksandra Piktus, Fabio Petroni, Vladimir
      Karpukhin, Naman Goyal, Heinrich K{\"u}ttler, Mike Lewis, Wen-tau Yih, Tim
      Rockt{\"a}schel, Sebastian Riedel, and Douwe Kiela.
    \newblock Retrieval-{{Augmented Generation}} for {{Knowledge-Intensive NLP
      Tasks}}, April 2021, {\em arXiv preprint arXiv:2005.11401}.
    
    \bibitem{paranjapeARTAutomaticMultistep2023}
    Bhargavi Paranjape, Scott Lundberg, Sameer Singh, Hannaneh Hajishirzi, Luke
      Zettlemoyer, and Marco~Tulio Ribeiro.
    \newblock {{ART}}: {{Automatic}} multi-step reasoning and tool-use for large
      language models, March 2023, {\em arXiv preprint arXiv:2303.09014}.
    
    \bibitem{weiChainofThoughtPromptingElicits2023}
    Jason Wei, Xuezhi Wang, Dale Schuurmans, Maarten Bosma, Brian Ichter, Fei Xia,
      Ed~Chi, Quoc Le, and Denny Zhou.
    \newblock Chain-of-{{Thought Prompting Elicits Reasoning}} in {{Large Language
      Models}}, January 2023, {\em arXiv preprint arXiv:2201.11903}.
    
    \bibitem{yaoTreeThoughtsDeliberate2023}
    Shunyu Yao, Dian Yu, Jeffrey Zhao, Izhak Shafran, Thomas~L. Griffiths, Yuan
      Cao, and Karthik Narasimhan.
    \newblock Tree of {{Thoughts}}: {{Deliberate Problem Solving}} with {{Large
      Language Models}}, December 2023, {\em arXiv preprint arXiv:2305.10601}.
    
    \bibitem{yaoReActSynergizingReasoning2023}
    Shunyu Yao, Jeffrey Zhao, Dian Yu, Nan Du, Izhak Shafran, Karthik Narasimhan,
      and Yuan Cao.
    \newblock {{ReAct}}: {{Synergizing Reasoning}} and {{Acting}} in {{Language
      Models}}, March 2023, {\em arXiv preprint arXiv:2210.03629}.
    
    \bibitem{baiQwenTechnicalReport2023}
    Jinze Bai, Shuai Bai, Yunfei Chu, Zeyu Cui, Kai Dang, Xiaodong Deng, Yang Fan,
      Wenbin Ge, Yu~Han, Fei Huang, Binyuan Hui, Luo Ji, Mei Li, Junyang Lin, Runji
      Lin, Dayiheng Liu, Gao Liu, Chengqiang Lu, Keming Lu, Jianxin Ma, Rui Men,
      Xingzhang Ren, Xuancheng Ren, Chuanqi Tan, Sinan Tan, Jianhong Tu, Peng Wang,
      Shijie Wang, Wei Wang, Shengguang Wu, Benfeng Xu, Jin Xu, An~Yang, Hao Yang,
      Jian Yang, Shusheng Yang, Yang Yao, Bowen Yu, Hongyi Yuan, Zheng Yuan,
      Jianwei Zhang, Xingxuan Zhang, Yichang Zhang, Zhenru Zhang, Chang Zhou,
      Jingren Zhou, Xiaohuan Zhou, and Tianhang Zhu.
    \newblock Qwen {{Technical Report}}, September 2023, {\em arXiv preprint arXiv:2309.16609}.
    
\end{thebibliography}





\newpage
\appendix

\section{Appendix}

\subsection{Prompts for LLMs} \label{sec:llm-prompts}

System prompt for agent1 under \texttt{ReAct} strategy:
\begin{lstlisting}
You are a helpful assistant like a pirate.
Perform the task to the best of your ability.
You is here to assist.
TOOLS:
------
You has access to the following tools:
calc_compressor_eff(inlet_T: float, inlet_P: float, outlet_T: float, outlet_P: float) -> dict - Calculate the isentropic efficiency of a COMPRESSOR by inputting inlet temperature and pressure, outlet temperature and pressure
calc_turbine_eff(inlet_T: float, inlet_P: float, outlet_T: float, outlet_P: float) -> dict - Calculate the isentropic efficiency of a TURBINE by inputting inlet temperature and pressure, outlet temperature and pressure
calc_nozzle(inlet_T: float, inlet_P: float, outlet_P: float, throat_area: float) -> dict - Calculate NOZZLE condition like the mass flow rate wether choked or not, by inputting the inlet temperature, inlet pressure, outlet pressure, and throat area
calc_burner_outlet(inlet_T: float, inlet_P: float, W_fuel: float, W_air: float) -> dict - Calculate the outlet temperature and pressure of a combustion chamber (BURNER) by inputting the inlet temperature, inlet pressure, fuel flow rate, and air flow rate
To use a tool, please use the following format:
```
Thought: Do I need to use a tool? Yes
Action: the action to take, should be one of [calc_compressor_eff, calc_turbine_eff, calc_nozzle, calc_burner_outlet]
Action Input: the input to the action
Observation: the result of the action
```
When you have a response to say to the Human, or if you do not need to use a tool, you MUST use the format:
```
Thought: Do I need to use a tool? No
Final Answer: [your response here]
```
You always MUST use the format:
```
Thought: you should always think about what to do
Action: the action to take, should be one of [calc_compressor_eff, calc_turbine_eff, calc_nozzle, calc_burner_outlet]
Action Input: the input to the action
Observation: the result of the action
... (this Thought/Action/Action Input/Observation can repeat N times)
Thought: I now know the final answer
Final Answer: the final answer to the original input question
```

You are familiar with the basic structure of a gas turbine, comprising a compressor, combustion chamber(burner), turbine, and nozzle. 
The components are connected in sequence, thus satisfying these relationships: 
* inlet-to-outlet correspondence: compressor outlet = burner inlet; burner outlet = turbine inlet; turbine outlet = nozzle inlet
* equality of flow rates across components: W_comp = W_burner = W_turbine = W_nozzle = W_air

When asked for a single component information, you can call the corresponding tool and answer directly

Please assist in implementing these calculations to analyze the performance of a gas turbine.


Your answer must always contain points:
(Thought, Final Answer) or (Thought, Action, Action Input)
Your answer can`t contain both points 'Final Answer' and 'Action'

You should use ONE tool at a time, think STEP BY STEP
Begin!
\end{lstlisting}

System prompt for agent2:
\begin{lstlisting}
You have access to the following tools:

[
  {
    "name": "calc_compressor_eff",
    "description": "calc_compressor_eff(inlet_T: float, inlet_P: float, outlet_T: float, outlet_P: float) -> dict - Calculate the isentropic efficiency of a COMPRESSOR by inputting inlet temperature and pressure, outlet temperature and pressure",
    "parameters": {
      "type": "object",
      "properties": {
        "inlet_T": {
          "type": "number"
        },
        "inlet_P": {
          "type": "number"
        },
        "outlet_T": {
          "type": "number"
        },
        "outlet_P": {
          "type": "number"
        }
      },
      "required": [
        "inlet_T",
        "inlet_P",
        "outlet_T",
        "outlet_P"
      ]
    }
  },
  {
    "name": "calc_turbine_eff",
    "description": "calc_turbine_eff(inlet_T: float, inlet_P: float, outlet_T: float, outlet_P: float) -> dict - Calculate the isentropic efficiency of a TURBINE by inputting inlet temperature and pressure, outlet temperature and pressure",
    "parameters": {
      "type": "object",
      "properties": {
        "inlet_T": {
          "type": "number"
        },
        "inlet_P": {
          "type": "number"
        },
        "outlet_T": {
          "type": "number"
        },
        "outlet_P": {
          "type": "number"
        }
      },
      "required": [
        "inlet_T",
        "inlet_P",
        "outlet_T",
        "outlet_P"
      ]
    }
  },
  {
    "name": "calc_nozzle",
    "description": "calc_nozzle(inlet_T: float, inlet_P: float, outlet_P: float, throat_area: float) -> dict - Calculate NOZZLE condition like the mass flow rate wether choked or not, by inputting the inlet temperature, inlet pressure, outlet pressure, and throat area",
    "parameters": {
      "type": "object",
      "properties": {
        "inlet_T": {
          "type": "number"
        },
        "inlet_P": {
          "type": "number"
        },
        "outlet_P": {
          "type": "number"
        },
        "throat_area": {
          "type": "number"
        }
      },
      "required": [
        "inlet_T",
        "inlet_P",
        "outlet_P",
        "throat_area"
      ]
    }
  },
  {
    "name": "calc_burner_outlet",
    "description": "calc_burner_outlet(inlet_T: float, inlet_P: float, W_fuel: float, W_air: float) -> dict - Calculate the outlet temperature and pressure of a combustion chamber (BURNER) by inputting the inlet temperature, inlet pressure, fuel flow rate, and air flow rate",
    "parameters": {
      "type": "object",
      "properties": {
        "inlet_T": {
          "type": "number"
        },
        "inlet_P": {
          "type": "number"
        },
        "W_fuel": {
          "type": "number"
        },
        "W_air": {
          "type": "number"
        }
      },
      "required": [
        "inlet_T",
        "inlet_P",
        "W_fuel",
        "W_air"
      ]
    }
  }
]

You must always select one of the above tools and respond with only a JSON object matching the following schema:

{
  "tool": <name of the selected tool>,
  "tool_input": <parameters for the selected tool, matching the tool's JSON schema>
}
\end{lstlisting}

\subsection{Tools for gas turbine components} \label{sec:tools-code}

\begin{lstlisting}[language=Python]
from langchain_core.tools import tool

gamma, R = 1.4, 287.052874

@tool
def calc_compressor_eff(inlet_T : float, inlet_P : float, outlet_T : float, outlet_P : float) -> dict:
    """Calculate the isentropic efficiency of a COMPRESSOR by inputting inlet temperature and pressure, outlet temperature and pressure"""
    return {"comp_isentropic_eff": ((outlet_P / inlet_P) ** ((gamma - 1) / gamma) - 1) / (outlet_T / inlet_T  - 1 )}

@tool
def calc_turbine_eff(inlet_T : float, inlet_P : float, outlet_T : float, outlet_P : float) -> dict:
    """Calculate the isentropic efficiency of a TURBINE by inputting inlet temperature and pressure, outlet temperature and pressure"""
    return {"turb_isentropic_eff": (1 - outlet_T / inlet_T) / (1 - (outlet_P / inlet_P) ** ((gamma - 1) / gamma))}

@tool
def calc_nozzle(inlet_T : float, inlet_P : float, outlet_P : float, throat_area : float) -> dict:
    """Calculate NOZZLE condition like the mass flow rate wether choked or not, by inputting the inlet temperature, inlet pressure, outlet pressure, and throat area"""
    nozz_pr = inlet_P / outlet_P
    nozz_mach = ((nozz_pr ** ((gamma - 1) / gamma) - 1) * (2 / (gamma - 1))) ** 0.5
    nozz_outlet_mach = min(nozz_mach, 1)
    nozz_w = inlet_P * throat_area / inlet_T**0.5 * gamma**0.5 * nozz_outlet_mach * (1+(gamma-1)/2*nozz_outlet_mach**2)**(-(gamma+1)/(2*gamma-2)) / R**0.5
    return {"W_nozz": nozz_w, "chocked": "yes" if nozz_mach > 1 else "no"}

@tool
def calc_burner_outlet(inlet_T : float, inlet_P : float, W_fuel : float, W_air : float) -> dict:
    """Calculate the outlet temperature and pressure of a combustion chamber (BURNER) by inputting the inlet temperature, inlet pressure, fuel flow rate, and air flow rate"""
    outlet_T = inlet_T + (48.6e6 / 1005) * W_fuel / W_air
    return {
        "burner_outlet_T": outlet_T,
        "burner_outlet_P": inlet_P * 0.95
    }
\end{lstlisting}

\subsection{Test questions for LLM agents} \label{sec:test-questions}

\begin{table}[H]
    \scriptsize
    \centering
    \begin{minipage}{0.95\linewidth}
        Questions of a single component can be solved with a single tool call
        \centering
        \ttfamily
        \begin{tabular}{p{0.12\linewidth} p{0.88\linewidth}}
            \toprule
            Question 1 & Please help me calculate the compressor efficiency with the following inlet conditions: temperature and pressure of 300K and 101325Pa, respectively; and outlet conditions: temperature and pressure of 420K and 201325Pa, respectively. \\
            Question 2 & Please assist me in calculating the efficiency of the turbine at the following inlet conditions: temperature and pressure of 1300K and 1601325Pa, respectively; and outlet conditions: temperature and pressure of 820K and 201325Pa, respectively. \\
            Question 3 & Please help me determine the nozzle flow rate and whether there is any blockage with the following conditions: inlet temperature and pressure of 420K and 201325Pa, respectively; outlet pressure of 101325Pa; and nozzle cross-sectional area of 0.25 m2. \\
            Question 4 & Please calculate the outlet conditions of the combustion chamber for me with the following inlet conditions: temperature and pressure of 1300K and 801325Pa, respectively; air flow rate of 85 kg/s; and fuel flow rate of 1.4 kg/s. \\
            Question 5 & Please help me calculate the flow rate through a nozzle with a minimum area of 0.3 m2, at atmospheric pressure of 101325Pa, and with inlet conditions of 420K and an inlet pressure of 400kPa. \\
            \bottomrule
        \end{tabular}
    \end{minipage}%
\end{table}

\begin{table}[H]
    \scriptsize
    \centering
    \begin{minipage}{0.95\linewidth}
        Questions that require chain thinking and continuous use of tools to solve
        \centering
        \ttfamily
        \begin{tabular}{p{0.12\linewidth} p{0.88\linewidth}}
            \toprule
            Question 6 &  I tested a gas turbine with the following data: Atmospheric temperature is 300K, pressure is 101325Pa. Compressor outlet / burner inlet temperature is 700K, pressure is 1800kPa, fuel flow rate of 1.5 kg/s. Nozzle inlet temperature of 620 K, pressure of 301325Pa, nozzle area is 4.24 m2. Please help me calculate the adiabatic efficiency of the compressor and the turbine. \\
            Question 7 & I tested a gas turbine with the following data: Atmospheric temperature is 300K, pressure is 101325Pa. Compressor outlet / burner inlet temperature is 700K, pressure is 1800kPa, fuel flow rate of 1.5 kg/s. Nozzle inlet temperature of 620 K, pressure of 301325Pa, nozzle area is 4.24 m2.  Please help me calculate the adiabatic efficiency of the compressor and the turbine. {\color{RoyalBlue} You should calculate nozzle first for air mass flow, and then calculate burner for turbine inlet. Then calculate the turbine efficiency.} \\
            \bottomrule
        \end{tabular}
    \end{minipage}%
\end{table}

\subsection{Response of LLM agents} \label{sec:agent-response}

\begin{table}[H]
    \centering
    \scriptsize
    \begin{minipage}{0.95\linewidth}
        Correct answers to questions 1 from the LLM agent built by llama3-70B
        \centering
        \ttfamily
        \begin{tabular}{p{0.12\linewidth} p{0.88\linewidth}}
            \toprule
            Agent1 & \\
            System &  System prompt for agent1 \\
            Human &  Please help me calculate the compressor efficiency with the following inlet conditions: temperature and pressure of 300K and 101325Pa, respectively; and outlet conditions: temperature and pressure of 420K and 201325Pa, respectively. AI scratchpad:  \\
            LLM & Thought: Do I need to use a tool? Yes. Action: calc\_compressror\_eff. Action Input: inlet\_T = 300, inlet\_P = 101325, outlet\_T = 420, outlet\_P = 201325. \\
            \midrule
            Agent2 & \\
            System &  System prompt for agent2 \\
            Human & Action: calc\_compressror\_eff. Action Input: inlet\_T = 300, inlet\_P = 101325, outlet\_T = 420, outlet\_P = 201325. \\
            LLM & \{ "tool": "calc\_compressor\_eff", "tool\_input": \{ "inlet\_T": 300, "inlet\_P": 101325, "outlet\_T": 420, "outlet\_P": 201325 \}\} \\
            \midrule
            Function & \{"comp\_isentropic\_eff": 0.5418276784464716\} \\
            \midrule
            Agent1 & \\
            System &  System prompt for agent1 \\
            Human &  Please help me calculate the compressor efficiency with the following inlet conditions: temperature and pressure of 300K and 101325Pa, respectively; and outlet conditions: temperature and pressure of 420K and 201325Pa, respectively. AI scratchpad: Thought: Do I need to use a tool? Yes. Action: calc\_compressror\_eff. Action Input: inlet\_T = 300, inlet\_P = 101325, outlet\_T = 420, outlet\_P = 201325. Observation: \{"comp\_isentropic\_eff": 0.5418276784464716\}. \\
            LLM & Thought: I now know the final answer. Final Answer: The isentropic efficiency of the compressor is approximately 54.18\%. \\
            \bottomrule
        \end{tabular}
    \end{minipage}%
\end{table}

\begin{table}[H]
    \centering
    \scriptsize
    \begin{minipage}{0.95\linewidth}
        Incorrect answers to questions 2 from the LLM agent built by llama3-8B
        \centering
        \ttfamily
        \begin{tabular}{p{0.12\linewidth} p{0.88\linewidth}}
            \toprule
            Agent1 & \\
            System &  System prompt for agent1 \\
            Human &  Please assist me in calculating the efficiency of the turbine at the following inlet conditions: temperature and pressure of 1300K and 1601325Pa, respectively; and outlet conditions: temperature and pressure of 820K and 201325Pa, respectively. AI scratchpad:  \\
            LLM & Thought: Do I need to use a tool? Yes. Action: calc\_turbine\_eff. Action Input: inlet\_T = 1300, inlet\_P = 1601325, outlet\_T = 820, outlet\_P = 201325. \\
            \midrule
            Agent2 & \\
            System &  System prompt for agent2 \\
            Human & Action: calc\_turbine\_eff. Action Input: inlet\_T = 1300, inlet\_P = 1601325, outlet\_T = 820, outlet\_P = 201325. \\
            LLM & \{ "turbine\_eff": 0.824312 \} \\
            \midrule
            Function & Error \\
            \bottomrule
        \end{tabular}
    \end{minipage}%
\end{table}

\begin{table}[H]
    \centering
    \scriptsize
    \begin{minipage}{0.95\linewidth}
        Incorrect answers to questions 2 from the LLM agent built by Qwen1.5-72B
        \centering
        \ttfamily
        \begin{tabular}{p{0.12\linewidth} p{0.88\linewidth}}
            \toprule
            Agent1 & \\
            System &  System prompt for agent1 \\
            Human &  Please assist me in calculating the efficiency of the turbine at the following inlet conditions: temperature and pressure of 1300K and 1601325Pa, respectively; and outlet conditions: temperature and pressure of 820K and 201325Pa, respectively. AI scratchpad:  \\
            LLM & Thought: Do I need to use a tool? Yes. Action: calc\_turbine\_eff. Action Input: inlet\_T = 1300, inlet\_P = 1601325, outlet\_T = 820, outlet\_P = 201325. \\
            \midrule
            Agent2 & \\
            System &  System prompt for agent2 \\
            Human & Action: calc\_turbine\_eff. Action Input: inlet\_T = 1300, inlet\_P = 1601325, outlet\_T = 820, outlet\_P = 201325. \\
            LLM & \{ "tool": "calc\_turbine\_eff", "tool\_input": \{ "inlet\_ T": 1300, "inlet\_ P": 1601325, "outlet\_ T": 820, "outlet\_ P": 201325 \}\} \\
            \midrule
            Function & Error \\
            \bottomrule
        \end{tabular}
    \end{minipage}%
\end{table}

\begin{table}[H]
    \centering
    \scriptsize
    \begin{minipage}{0.95\linewidth}
        Correct answers to questions 2 from the LLM agent built by llama3-70B (Outputs of agent2 and function calls are omitted)
        \centering
        \ttfamily
        \begin{tabular}{p{0.12\linewidth} p{0.88\linewidth}}
            \toprule
            Agent1 & \\
            System &  System prompt for agent1 \\
            Human &  Please assist me in calculating the efficiency of the turbine at the following inlet conditions: temperature and pressure of 1300K and 1601325Pa, respectively; and outlet conditions: temperature and pressure of 820K and 201325Pa, respectively. AI scratchpad:  \\
            LLM & Thought: Do I need to use a tool? Yes. Action: calc\_turbine\_eff. Action Input: inlet\_T = 1300, inlet\_P = 1601325, outlet\_T = 820, outlet\_P = 201325. \\
            \midrule
            ... & ...\\
            \midrule
            Human & ... Observation: \{"turb\_isentropic\_eff": 0.8259391320308387\}. \\
            LLM & Thought: I now know the final answer. Final Answer: The isentropic efficiency of the turbine is approximately 82.59\%. \\
            \bottomrule
        \end{tabular}
    \end{minipage}%
\end{table}

\begin{table}[H]
    \centering
    \scriptsize
    \begin{minipage}{0.95\linewidth}
        Correct answers to questions 3 from the LLM agent built by llama3-70B (Outputs of agent2 and function calls are omitted)
        \centering
        \ttfamily
        \begin{tabular}{p{0.12\linewidth} p{0.88\linewidth}}
            \toprule
            Agent1 & \\
            System &  System prompt for agent1 \\
            Human &  Please help me determine the nozzle flow rate and whether there is any blockage with the following conditions: inlet temperature and pressure of 420K and 201325Pa, respectively; outlet pressure of 101325Pa; and nozzle cross-sectional area of 0.25 m2. AI scratchpad:  \\
            LLM & Thought: Do I need to use a tool? Yes. Action: calc\_nozzle. Action Input: inlet\_T = 420, inlet\_P = 201325, outlet\_P = 101325, throat\_area = 0.25. \\
            \midrule
            ... & ...\\
            \midrule
            Human & ... Observation: \{"W\_nozz": 99.25500171944374, "chocked": "yes"\}. \\
            LLM & Thought: I now know the final answer. Final Answer: The nozzle is choked and has a mass flow rate of approximately 99.26 kg/s. \\
            \bottomrule
        \end{tabular}
    \end{minipage}%
\end{table}

\begin{table}[H]
    \centering
    \scriptsize
    \begin{minipage}{0.95\linewidth}
        Correct answers to questions 4 from the LLM agent built by llama3-70B (Outputs of agent2 and function calls are omitted)
        \centering
        \ttfamily
        \begin{tabular}{p{0.12\linewidth} p{0.88\linewidth}}
            \toprule
            Agent1 & \\
            System &  System prompt for agent1 \\
            Human &  Please calculate the outlet conditions of the combustion chamber for me with the following inlet conditions: temperature and pressure of 1300K and 801325Pa, respectively; air flow rate of 85 kg/s; and fuel flow rate of 1.4 kg/s. AI scratchpad:  \\
            LLM & Thought: Do I need to use a tool? Yes. Action: calc\_burner\_outlet. Action Input: inlet\_T = 1300, inlet\_P = 801325, W\_air = 85, W\_fuel = 1.4. \\
            \midrule
            ... & ...\\
            \midrule
            Human & ... Observation: \{"burner\_outlet\_T": 2096.488147497805, "burner\_outlet\_P": 761258.75\}. \\
            LLM & Thought: I now know the burner outlet condition. Final Answer: The outlet temperature and pressure of the combustion chamber (BURNER) are approximately 2096.49 K and 761258.75 Pa, respectively. \\
            \bottomrule
        \end{tabular}
    \end{minipage}%
\end{table}

\begin{table}[H]
    \centering
    \scriptsize
    \begin{minipage}{0.95\linewidth}
        Correct answers to questions 5 from the LLM agent built by llama3-70B (Outputs of agent2 and function calls are omitted)
        \centering
        \ttfamily
        \begin{tabular}{p{0.12\linewidth} p{0.88\linewidth}}
            \toprule
            Agent1 & \\
            System &  System prompt for agent1 \\
            Human &  Please help me calculate the flow rate through a nozzle with a minimum area of 0.3 m2, at atmospheric pressure of 101325Pa, and with inlet conditions of 420K and an inlet pressure of 400kPa. AI scratchpad:  \\
            LLM & Thought: Do I need to use a tool? Yes. Action: calc\_nozzle. Action Input: inlet\_T = 420, inlet\_P = 400000, outlet\_P = 101325, throat\_area = 0.3. \\
            \midrule
            ... & ...\\
            \midrule
            Human & ... Observation: \{"W\_nozz": 236.64423606274923, "chocked": "yes"\}. \\
            LLM & Thought: I now know the final answer. Final Answer: The mass flow rate of the nozzle is approximately 236.64 kg/s and it is choked. \\
            \bottomrule
        \end{tabular}
    \end{minipage}%
\end{table}

\begin{table}[H]
    \centering
    \scriptsize
    \begin{minipage}{0.95\linewidth}
        Incorrect answers to questions 6 from the LLM agent built by llama3-70B (Outputs of agent2 and function calls are omitted)
        \centering
        \ttfamily
        \begin{tabular}{p{0.12\linewidth} p{0.88\linewidth}}
            \toprule
            Agent1 & \\
            System &  System prompt for agent1 \\
            Human &  I tested a gas turbine with the following data: Atmospheric temperature is 300K, pressure is 101325Pa. Compressor outlet / burner inlet temperature is 700K, pressure is 1800kPa, fuel flow rate of 1.5 kg/s. Nozzle inlet temperature of 620 K, pressure of 301325Pa, nozzle area is 4.24 m2. Please help me calculate the adiabatic efficiency of the compressor and the turbine. AI scratchpad:  \\
            LLM & Thought: Do I need to use a tool? Yes. Action: calc\_compressor\_eff. Action Input: inlet\_T = 300, inlet\_P = 101325, outlet\_T = 700, outlet\_P = 1800000. \\
            \midrule
            ... & ...\\
            \midrule
            Human & ... Observation: \{"comp\_isentropic\_eff": 0.9563858170355589\}. \\
            LLM & Thought: I need to calculate the adiabatic efficiency of the turbine. Action: calc\_turbine\_eff. Action Input: inlet\_T = 700, inlet\_P = 1800000, outlet\_T = 620, outlet\_P = 101325. \\
            \midrule
            ... & ...\\
            \midrule
            Human & I tested a gas turbine with the following data: Atmospheric temperature is 300K, pressure is 101325Pa. Compressor outlet / burner inlet temperature is 700K, pressure is 1800kPa, fuel flow rate of 1.5 kg/s. Nozzle inlet temperature of 620 K, pressure of 301325Pa, nozzle area is 4.24 m2. Please help me calculate the adiabatic efficiency of the compressor and the turbine. AI scratchpad: Do I need to use a tool? Yes. Action: calc\_compressor\_eff. Action Input: inlet\_T = 300, inlet\_P = 101325, outlet\_T = 700, outlet\_P = 1800000. Observation: \{"comp\_isentropic\_eff": 0.9563858170355589\}. Thought: I need to calculate the adiabatic efficiency of the turbine. Action: calc\_turbine\_eff. Action Input: inlet\_T = 700, inlet\_P = 1800000, outlet\_T = 620, outlet\_P = 101325. Observation: \{"turb\_isentropic\_eff": 0.20390883937551146\}. \\
            LLM & Thought: I now know the final answer. Final Answer: The adiabatic efficiency of the compressor is approximately 95.64\% and the adiabatic efficiency of the turbine is approximately 20.39\%. \\
            \bottomrule
        \end{tabular}
    \end{minipage}%
\end{table}

\begin{table}[H]
    \centering
    \scriptsize
    \begin{minipage}{0.95\linewidth}
        Correct answers to questions 7 from the LLM agent built by llama3-70B (Outputs of agent2 and function calls are omitted)
        \centering
        \ttfamily
        \begin{tabular}{p{0.12\linewidth} p{0.88\linewidth}}
            \toprule
            Agent1 & \\
            System &  System prompt for agent1 \\
            Human &  I tested a gas turbine with the following data: Atmospheric temperature is 300K, pressure is 101325Pa. Compressor outlet / burner inlet temperature is 700K, pressure is 1800kPa, fuel flow rate of 1.5 kg/s. Nozzle inlet temperature of 620 K, pressure of 301325Pa, nozzle area is 4.24 m2.  Please help me calculate the adiabatic efficiency of the compressor and the turbine. {\color{RoyalBlue} You should calculate nozzle first for air mass flow, and then calculate burner for turbine inlet. Then calculate the turbine efficiency.} AI scratchpad:  \\
            LLM & Thought: Do I need to use a tool? Yes. Action: calc\_nozzle. Action Input: inlet\_T = 620, inlet\_P = 301325, outlet\_P = 101325, throat\_area = 4.24. \\
            \midrule
            ... & ...\\
            \midrule
            Human & ... Observation: \{"W\_nozz": 2073.693216788111, "chocked": "yes"\}. \\
            LLM & Thought: I need to use another tool to calculate the air mass flow rate at the burner. Action: calc\_burner\_outlet. Action Input: inlet\_T = 700, inlet\_P = 1800000, W\_fuel = 1.5, W\_air = 2073.693216788111. \\
            \midrule
            ... & ...\\
            \midrule
            Human & ... Observation: \{"burner\_outlet\_T": 734.9797708000352, "burner\_outlet\_P": 1710000.0\}. \\
            LLM & Thought: Do I need to use a tool? Yes. Action: calc\_turbine\_eff. Action Input: inlet\_T = 734.9797708000352, inlet\_P = 1710000.0, outlet\_T = 620, outlet\_P = 301325. \\
            \midrule
            ... & ...\\
            \midrule
            Human & ... Observation: \{"turb\_isentropic\_eff": 0.4000508940630961\}. \\
            LLM & Thought: Do I need to use a tool? Yes. Action: calc\_compressor\_eff. Action Input: inlet\_T = 300, inlet\_P = 101325, outlet\_T = 700, outlet\_P = 1800000. \\
            \midrule
            ... & ...\\
            \midrule
            Human & ... Observation: \{"comp\_isentropic\_eff": 0.9563858170355589\}.\\
            LLM & Thought: Thought: I now know the final answer. Final Answer: The adiabatic efficiency of the compressor is 0.9563858170355589 and the turbine is 0.4000508940630961. \\
            \bottomrule
        \end{tabular}
    \end{minipage}%
\end{table}

\end{document}